\pdfoutput=1

\documentclass[11pt]{article}

\usepackage[final]{acl}

\usepackage{times}
\usepackage{latexsym}

\usepackage[T1]{fontenc}

\usepackage[utf8]{inputenc}

\usepackage{microtype}

\usepackage{inconsolata}

\usepackage{graphicx}

\usepackage{amsmath}
\usepackage{amssymb}
\usepackage{multirow}
\usepackage{hyperref}
\usepackage{kotex}
\usepackage[normalem]{ulem}
\useunder{\uline}{\ul}{}
\usepackage{color, colortbl}
\usepackage{booktabs}
\usepackage{graphicx}
\usepackage{adjustbox}
\usepackage{subcaption}
\usepackage{authblk}

\definecolor{chembrayblue}{rgb}{0.6196, 0.7059 0.8275}
\definecolor{lightblue}{rgb}{0.886, 0.929, 0.996}

\title{Zero-shot Commonsense Reasoning over Machine Imagination}

\author{\textbf{Hyuntae Park\textsuperscript{1}}\thanks{\ \ These authors contributed equally to this work.}, \: 
  \textbf{Yeachan Kim\textsuperscript{1*}}, \:
  \textbf{Jun-Hyung Park\textsuperscript{2}}, \:
  \textbf{SangKeun Lee\textsuperscript{1,3}} \\
  \textsuperscript{1}Department of Artificial Intelligence, Korea University, Seoul, Republic of Korea \\
  \textsuperscript{2}Division of Language \& AI, Hankuk University of Foreign Studies, Seoul, Republic of Korea \\
  \textsuperscript{3}Department of Computer Science and Engineering, Korea University, Seoul, Republic of Korea \\
  \texttt{\{pht0639, yeachan, yalphy\}@korea.ac.kr, \ jhp@hufs.ac.kr} \\
}

\begin{document}
\maketitle
\begin{abstract}
Recent approaches to zero-shot commonsense reasoning have enabled Pre-trained Language Models (PLMs) to learn a broad range of commonsense knowledge without being tailored to specific situations. 
However, they often suffer from human reporting bias inherent in textual commonsense knowledge, leading to discrepancies in understanding between PLMs and humans. 
In this work, we aim to bridge this gap by introducing an additional information channel to PLMs. 
We propose \textsc{Imagine} ({Machine \textbf{Imagin}ation-based R\textbf{e}asoning}), a novel zero-shot commonsense reasoning framework designed to complement textual inputs with visual signals derived from machine-generated images. 
To achieve this, we enhance PLMs with imagination capabilities by incorporating an image generator into the reasoning process.
To guide PLMs in effectively leveraging machine imagination, we create a synthetic pre-training dataset that simulates visual question-answering. 
Our extensive experiments on diverse reasoning benchmarks and analysis show that \textsc{Imagine} outperforms existing methods by a large margin, highlighting the strength of machine imagination in mitigating reporting bias and enhancing generalization capabilities\footnote{Our code and data are available at \url{https://github.com/Park-ing-lot/Imagine}}.
\end{abstract}

\begin{figure}[ht!]
\centering
  \includegraphics[width=\linewidth]{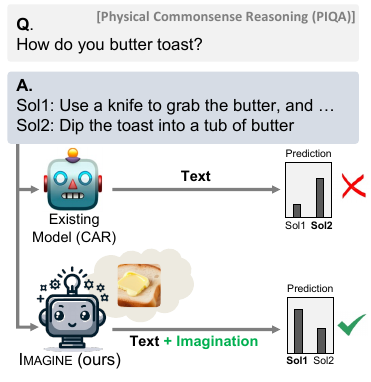}
    \caption{Example from the PIQA \cite{DBLP:conf/aaai/BiskZLGC20-piqa} with model predictions. Compared to the existing methods, \textsc{Imagine} performs reasoning with imagination.}
    \label{fig:fig1}
\end{figure}

\section{Introduction}
Commonsense reasoning has been considered a crucial milestone in the pursuit of artificial general intelligence \citep{gunning2018machine}. While Pre-trained Language Models (PLMs; \citealp{DBLP:conf/naacl/DevlinCLT19-bert, DBLP:conf/nips/BrownMRSKDNSSAA20-gpt}) often exhibit near-human reasoning capabilities after being fine-tuned on specific commonsense datasets, they face challenges in zero-shot scenarios where examples differ significantly from their training data distribution \citep{DBLP:journals/corr/abs-1909-08855-mitra, DBLP:conf/naacl/KimKKAHY22}. Overcoming this limitation is crucial for achieving human-level proficiency in natural language understanding.

One promising approach to this limitation is injecting commonsense knowledge from external Knowledge Bases (KBs; \citealp{DBLP:conf/aaai/SapBABLRRSC19-atomic, DBLP:journals/corr/abs-2206-01532-abstractatomic}) into PLMs. Specifically, this involves transforming knowledge entities into a question-answering (QA) format, resulting in a synthetic QA dataset. This constructed dataset is then used to train PLMs similarly to the pre-training phase. Since the knowledge bases can cover a wide spectrum of commonsense knowledge, this approach leads to substantial improvements in reasoning ability across diverse situations without specializing in specific knowledge \cite{DBLP:conf/emnlp/WangF0XLSB23-car,DBLP:journals/corr/abs-2401-07286-candle}.

However, they often suffer from human reporting bias \cite{DBLP:conf/cikm/GordonD13-reporting}, as textual commonsense knowledge only captures the most frequently occurring scenarios, thereby neglecting less common but equally critical knowledge necessary for comprehensive reasoning. Figure \ref{fig:fig1} illustrates a case where a recent model \cite{DBLP:conf/emnlp/WangF0XLSB23-car} fails to accurately reason about the question "\textit{How do you butter toast?}". Since the existing models rely solely on textual inputs, they often neglect contextual details, such as the fact that butter is typically too solid to be dipped. In contrast, humans can easily answer such questions by visually imagining the shape, solidity, and interactions of butter with other objects. 
This observation motivates us to explore additional modalities to complement textual commonsense knowledge.

In this paper, we introduce \textsc{Imagine} ({Machine \textbf{Imagin}ation-based R\textbf{e}asoning}), a novel zero-shot commonsense reasoning framework designed to circumvent the reporting bias inherent in textual inputs. Inspired by the cognitive studies highlighting the beneficial effects of visual imagery on language understanding \citep{Gambrell1986MentalIA, Dessalegn2013InteractionBL}, \textsc{Imagine} is designed to leverage visual signals to complement textual inputs. To achieve this, we integrate PLMs with a conditional image generator, enabling machine imagination capabilities. To guide the model in learning to utilize visual and textual inputs jointly, we create a Synthetic VQA dataset, which is then used to optimize PLMs. By acquiring a broad spectrum of commonsense knowledge along with visual signals, \textsc{Imagine} enhances reasoning capabilities while circumventing human reporting bias.

To verify the effectiveness of \textsc{Imagine}, we perform extensive experiments, encompassing diverse reasoning benchmarks, architectures, and scales. The experimental results convincingly demonstrate that \textsc{Imagine} surpasses existing methods, including large language models, in reasoning capabilities. Moreover, our in-depth analysis reveals that \textsc{Imagine} effectively enables PLMs to adaptively leverage machine imagination capabilities in a beneficial manner. The contributions of this paper include the following:
\begin{itemize}

    \item We introduce \textsc{Imagine}, a novel zero-shot commonsense reasoning framework, aimed at mitigating reporting bias and enhancing the generalizability of PLMs.
    
    \item We construct a Synthetic VQA dataset to enable PLMs to jointly utilize textual and visual signals while achieving commonsense reasoning ability. 

    \item We demonstrate that \textsc{Imagine} surpasses state-of-the-art zero-shot reasoning models across diverse reasoning tasks, highlighting the significance of machine imagination.
    
\end{itemize}

\begin{figure*}[t]
\centering
\subfloat[Construction procedures of Synthetic VQA dataset]{%
  \includegraphics[width=\linewidth]{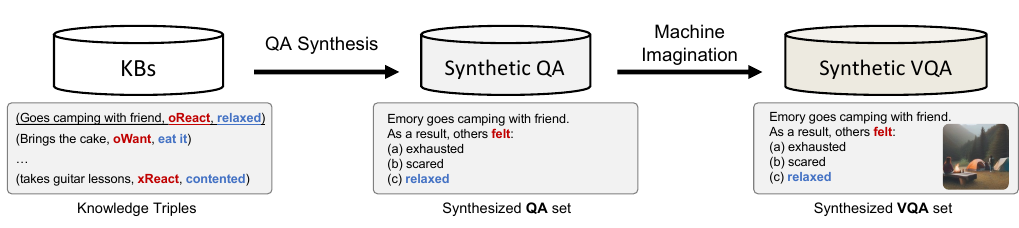}%
}

\subfloat[Inference and optimization procedures of \textsc{Imagine} (ours)]{%
  \includegraphics[width=\linewidth]{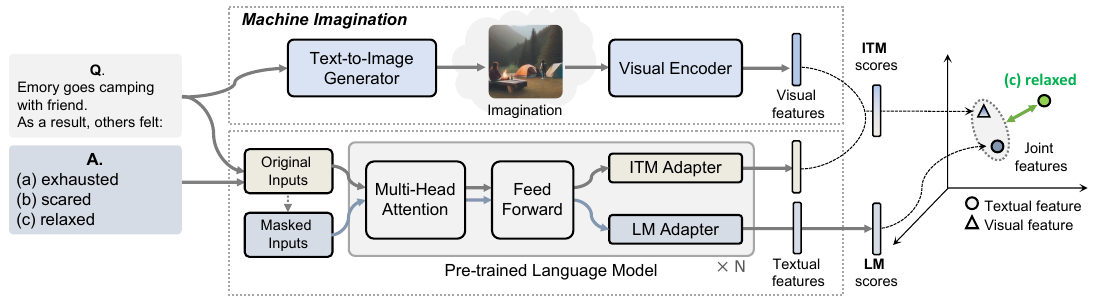}%
}
\caption{Overall procedures for (a) constructing a Synthetic VQA dataset and (b) the inference/optimization phase of \textsc{Imagine} (ours) using the given QA pair. The process starts with the textual pair consisting of a question and its answers, followed by the generation of visual signals (i.e., imagination) conditioned on the question. The two distinct features from visual and textual models are then utilized to derive a comprehensive prediction.}
\label{fig:model}
\end{figure*}

\section{Related Work}

\subsection{Zero-shot Commonsense Reasoning}

There are two major approaches to zero-shot commonsense reasoning. The first approach involves utilizing the inherent capabilities of the off-the-shelf PLMs without updating their parameters. For example, \citet{DBLP:journals/corr/abs-1806-02847} utilized the perplexity of vanilla language modeling, and \citet{DBLP:conf/emnlp/LiKHdBN22} leveraged PLMs with specifically-designed prompting. \citet{DBLP:conf/emnlp/ShwartzWBBC20-selftalk} solicited the commonsense knowledge from the language models through an iterative self-talk. Similarly, \citet{DBLP:conf/aaai/DouP22} obtained additional knowledge for reasoning based on the cloze-style translation. The second approach involves leveraging external commonsense knowledge bases (e.g., ATOMIC \cite{DBLP:conf/aaai/SapBABLRRSC19-atomic}, ConceptNet \cite{DBLP:conf/aaai/SpeerCH17-conceptnet}) to provide language models with additional knowledge. Specifically, recent studies have transformed the knowledge entities (e.g., triplets of (head, relation, tail)) into synthetic QA pairs and trained the models with them \cite{DBLP:conf/emnlp/BanerjeeB20-smlm,DBLP:conf/aaai/MaIFBNO21-ma}. Recently, \citet{DBLP:conf/emnlp/WangF0XLSB23-car} further improved the synthetic signals through a conceptualization process \citep{song2011short} which abstracts a commonsense knowledge triplet to many higher-level instances. Subsequently, \citet{DBLP:journals/corr/abs-2401-07286-candle} injected the instantiation phase into the process of synthetic dataset generation with the help of the generation capabilities of LLMs.


\subsection{Visual Information for Natural Language Understanding}

A few previous works have leveraged machine imagination to address Natural Language Understanding (NLU) problems. For example, \citet{tan2020vokenization} proposed VOKEN, which introduces visual supervision into language model pre-training by incorporating external knowledge from images retrieved for the tokens. Instead of retrieving visual information, \citet{lu2022imagination} proposed generating synthetic images (i.e., imagination) based on a generative model to tackle downstream NLU tasks. In the context of commonsense reasoning, \citet{liu2022things} utilized visual information to comprehend spatial commonsense knowledge (e.g., \textit{how big is a lion?}). Similar to the proposed method, \citet{DBLP:conf/emnlp/YangYZWYC22-zlavi} introduced Z-LaVI, which integrated visual information with PLMs through both retrieval and synthesis to achieve zero-shot reasoning abilities. Unlike previous approaches that employ visual signals directly, we introduce a distinct pre-training phase which allows the model to effectively utilize visual imagination for zero-shot reasoning.

\section{Machine Imagination-based Reasoning}

\begin{figure*}[ht!]
\centering
  \includegraphics[width=\linewidth]{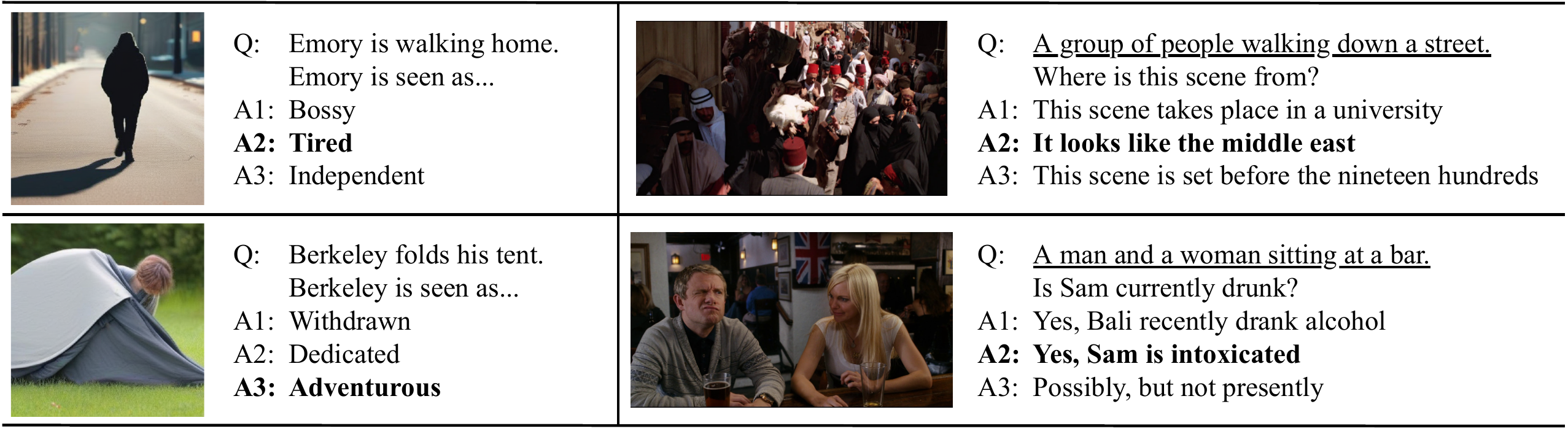}
    \caption{Examples of the Synthetic VQA dataset. The examples on the left are sourced from AbstractATOMIC \citep{DBLP:conf/emnlp/WangF0XLSB23-car}, while the two examples on the right are sourced from VCR \citep{DBLP:conf/cvpr/ZellersBFC19-vcr}. \textbf{Bold} indicates the correct answer, and {\ul underline} denotes the generated image caption.}
    \label{fig:vqa_examples}
\end{figure*}

In this section, we elaborate on the proposed method, namely \textsc{Imagine} (Machine \textbf{Imagin}ation-based R\textbf{e}asoning), for zero-shot commonsense reasoning.  The core strategy is to complement textual commonsense knowledge with visual signals derived from machine-generated images. To achieve this, we first couple the PLMs with a text-to-image generator (\S\ref{subsec:machine_imagination}), enabling machine imagination in text-based PLMs. We then construct a large-scale Synthetic VQA dataset to learn the joint use of textual and visual signals in the reasoning process (\S\ref{subsec:vqa}). By optimizing the model with additional signals that encapsulate commonsense knowledge, \textsc{Imagine} can effectively perform commonsense reasoning while avoiding human reporting bias inherent in textual inputs (\S\ref{subsec:injection}, \S\ref{subsec:inference}). The overall procedure is depicted in Figure \ref{fig:model}.

\subsection{Machine Imagination in PLMs}\label{subsec:machine_imagination}

We start by introducing the machine imagination in text-based PLMs. 
We denote PLMs as $\mathcal{M}_{T}$, which serve as the backbone for zero-shot commonsense reasoning.
For machine imagination, we incorporate two additional models to process visual signals. Specifically, we introduce: (i) a text-to-image generator, $\mathcal{M}_{T2I}$, which creates relevant images by conditioning the textual inputs, and (ii) a visual encoder, $\mathcal{M}_{I}$, which acts as a feature extractor for the given images.

The overall mechanism of machine imagination operates as follows: Given a textual input, the text-to-image model $\mathcal{M}_{T2I}$ initially generates an image that captures the essence of the text. With these generated images linked to textual inputs, both PLMs, $\mathcal{M}_{T}$, and the visual encoder, $\mathcal{M}_{I}$, jointly encode the textual input and the generated image. The resultant features are then utilized to derive the comprehensive predictions.

\subsection{Synthetic VQA Construction}\label{subsec:vqa}

Following the previous works \cite{DBLP:conf/aaai/MaIFBNO21-ma,DBLP:conf/emnlp/WangF0XLSB23-car}, we achieve zero-shot commonsense reasoning ability by constructing the synthetic QA dataset from the knowledge base. On top of this dataset, we build a synthetic visual question-answering (Synthetic VQA) dataset with the help of machine imagination. Additionally, we incorporate a visual commonsense dataset that contains real images \citep{DBLP:conf/cvpr/ZellersBFC19-vcr}. The dataset is designed to: (i) instill commonsense reasoning abilities in PLMs and (ii) teach them to harmoniously utilize both textual and visual inputs. Examples of the Synthetic VQA dataset can be found in Figure \ref{fig:vqa_examples}.

The objective of this process is to construct VQA pairs $(Q, A, I)$, where each pair includes a natural language question $Q$, a set of $n$ answer choices $A={A_1, A_2, ..., A_n}$, including one ground-truth answer and $n-1$ distractors, along with an image $I$ that corresponds to the question.

\paragraph{Synthetic QA} We first construct textual QA pairs from the KBs by following the recent work \cite{DBLP:conf/emnlp/WangF0XLSB23-car}. Specifically, we transform the knowledge entities into the QA pairs through the conceptualized augmentation of the entities \cite{DBLP:conf/emnlp/WangF0XLSB23-car} with the pre-defined natural language templates (e.g., the relation of \textit{xWant} is transformed to \textit{As a result, PersonX wanted to}). This process results in textual synthetic QA pairs $(Q, A)$.

\paragraph{Synthetic VQA} On the textual synthetic QA pairs, we input the textual question $Q$ to the text-to-image model $\mathcal{M}_{T2I}$ to generate the visual counterpart $I$ that depicts the scenarios described in each question. These generated images provide an additional layer of information, offering a visual context that enhances the reasoning ability based not only on textual descriptions but also on visual evidence. This augmentation leverages the strengths of visual imagery on language understanding \cite{Gambrell1986MentalIA,Dessalegn2013InteractionBL}, potentially improving the robustness and accuracy of the model predictions.

However, relying solely on the synthetic relationships between QA pairs and generated images can introduce challenges related to the alignment of visual content since machines often fail to generate well-aligned images with textual inputs \citep{DBLP:conf/iclr/FengHFJANBWW23-t2i_alignment}.
Therefore, we augment the Synthetic VQA pairs with the widely used Visual Commonsense Reasoning (VCR) dataset \cite{DBLP:conf/cvpr/ZellersBFC19-vcr}. Each pair from this dataset consists of $(Q, A, R, I)$, where $R$ is a rationale for the correct answer; however, we omit $R$ since our focus is on the QA pairs associated with relevant images. Additionally, to enrich the input and enhance visual comprehension for PLMs, we generate textual context information for each image using an image captioning model\footnote{We use InstructBLIP \citep{DBLP:conf/nips/Dai0LTZW0FH23-instructblip} for captioning.}, which we prepend as a prefix to each $Q$\footnote{More details of Synthetic VQA are in Appendix \ref{app:data}.}. 

\subsection{Pre-training \textsc{Imagine} on Synthetic VQA}
\label{subsec:injection}
Based on the Synthetic VQA dataset, we integrate commonsense knowledge into the models. Since \textsc{Imagine} involves two distinct modalities (i.e., text and image), we introduce two separate objectives to select the best answer choice: Language Modeling (LM) and Image-Text Matching (ITM). To obtain the LM scores, we calculate the masked language modeling loss for the Transformer encoder-based model, formulated as:
\begin{equation}
S_{LM}(T) = -\frac{1}{m} \sum_{t=1}^{m} \log P(w_t | ...w_{t-1}, w_{t+1}...).
\end{equation}
For the decoder-based model, we compute the auto-regressive language modeling loss, defined as:
\begin{equation}
S_{LM}(T) = -\frac{1}{m} \sum_{t=1}^{m} \log P(w_t | w_{1}...w_{t-1}),
\end{equation}
where $w_i$ denotes the $i$-th word, and $m$ is the number of tokens in the sequence $T$. 
To compute the ITM scores, we first contextualize the visual features based on the textual sequences. Let the visual features from the visual encoder $\mathcal{M}_{I}$ be denoted as $V$, we derive the contextualized visual features as follows:
\begin{equation}
    C = \text{softmax}(\frac{\vec{T}V^{\top}}{\sqrt{d_v}})V,
\end{equation}
where $\vec{T}$ is the feature vector from the PLMs $\mathcal{M}_{T}$. 
For the encoder-based model, we use the final hidden state of the $[\text{CLS}]$ token as the context vector, and for the decoder-based model, we use the hidden state of the last token as the context vector.\
$d_v$ is the dimension of visual features. We then achieve the ITM scores by calculating the similarity between contextualized visual features and textual features as follows:
\begin{equation}
    S_{I}(T, V) = \text{sim}(\vec{T}, C),
\end{equation}
where $\text{sim}(\cdot)$ denotes the cosine similarity function. By combining two different scores, we produce the joint scores $S_{J}$ as follows:
\begin{equation}
    S_{J}(T, V) = \frac{1}{2}(S_{LM}(T) + S_{I}(T, V)),
\end{equation}
After calculating all scores $S^{(1)}, S^{(2)}, ..., S^{(n)}$ for $n$ answer candidates, we calculate the marginal ranking loss defined as:
\begin{equation}
    \mathcal{L}_{QA}(S) = \frac{1}{n}\sum_{i=1, i \neq y}^{n} \max(0, \eta -S^{(y)} + S^{(i)}),
\end{equation}
where $y$ indicates the index of the correct answer and $\eta$ is the pre-defined margin. The overall objectives are as follows:
\begin{equation}
    \mathcal{L} = \mathcal{L}_{QA}(S_{LM}) + \mathcal{L}_{QA}(S_I) + \mathcal{L}_{QA}(S_J).   
\end{equation}
However, we have empirically observed that the ITM objective prevents the model from learning the LM objective, which is essential for developing reasoning capabilities. To mitigate the conflict between these two objectives, we introduce two distinct adapters \cite{DBLP:conf/iclr/HeZMBN22-parallel_adapter}, LM adapter and ITM adapter. Each adapter is trained separately with a different focus. It is important to note that only the weights within these adapters are optimized during training; all other parameters remain frozen. By separating the parameters for objectives, we can effectively reduce conflicts between them.

\subsection{Inference from $\textsc{Imagine}$}
\label{subsec:inference}
For the zero-shot evaluation, we use the same strategy to compute the LM and ITM scores after synthesizing the image based on the question.
Then we assemble two scores to derive the model's prediction after obtaining the probability distribution through softmax.
\begin{equation}
    P(S) = \text{softmax}(S^{(1)}, S^{(2)}, ..., S^{(n)}),
\end{equation}
\begin{equation}
    P(A|Q) = (1-\lambda) \cdot P(S_M) + \lambda \cdot P(S_I),
\end{equation}
where $\lambda$ is an ensemble coefficient that controls the contributions between textual and visual features.

\begin{table*}[ht!]
\footnotesize
\centering
\resizebox{\textwidth}{!}{%
\begin{tabular}{@{}lccccccc@{}}
\toprule
\multicolumn{1}{l|}{Method} & \multicolumn{1}{c|}{KB} & $\alpha$NLI & CSQA & PIQA & SIQA & \multicolumn{1}{c|}{WG} & Avg. \\ \midrule
\multicolumn{1}{l|}{GPT-2-L \citep{radford2019language-gpt2}} & \multicolumn{1}{c|}{-} & 56.5 & 41.4 & 68.9 & 44.6 & \multicolumn{1}{c|}{53.2} & 52.9 \\
\multicolumn{1}{l|}{RoBERTa-L \citep{DBLP:journals/corr/abs-1907-11692-roberta}} & \multicolumn{1}{c|}{-} & 65.6 & 45.0 & 67.6 & 47.3 & \multicolumn{1}{c|}{57.5} & 56.6 \\
\multicolumn{1}{l|}{DeBERTa-v3-L \citep{DBLP:conf/iclr/HeGC23-debertav3}} & \multicolumn{1}{c|}{-} & 59.9 & 25.4 & 44.8 & 47.8 & \multicolumn{1}{c|}{50.3} & 45.6 \\ \midrule
\multicolumn{1}{l|}{RoBERTa-L (MR; \citealp{DBLP:conf/aaai/MaIFBNO21-ma})} & \multicolumn{1}{c|}{AT} & 70.8 & 64.2 & 72.1 & 63.1 & \multicolumn{1}{c|}{59.6} & 66.0 \\
\multicolumn{1}{l|}{Zero-shot Fusion \citep{DBLP:conf/naacl/KimKKAHY22}} & \multicolumn{1}{c|}{AT, CN, WD, WN} & 72.5 & 68.2 & 72.9 & \textbf{66.6} & \multicolumn{1}{c|}{60.8} & 68.2 \\
\multicolumn{1}{l|}{CAR-RoBERTa-L \citep{DBLP:conf/emnlp/WangF0XLSB23-car}} & \multicolumn{1}{c|}{AbsAT} & 72.7 & 66.3 & 73.2 & 64.0 & \multicolumn{1}{c|}{62.0} & 67.6 \\
\multicolumn{1}{l|}{CAR-DeBERTa-v3-L \citep{DBLP:conf/emnlp/WangF0XLSB23-car}} & \multicolumn{1}{c|}{AbsAT} & 79.6 & 69.3 & 78.6 & 64.0 & \multicolumn{1}{c|}{{\ul 78.2}} & 73.9 \\
\multicolumn{1}{l|}{CANDLE-DeBERTa-v3-L \citep{DBLP:journals/corr/abs-2401-07286-candle}} & \multicolumn{1}{c|}{CANDLE} & {\ul 81.2} & {\ul 69.9} & {\ul 80.3} & 65.9 & \multicolumn{1}{c|}{\textbf{78.3}} & {\ul 75.1} \\ 
\multicolumn{1}{l|}{CANDLE-VERA-T5-xxl \citep{DBLP:journals/corr/abs-2401-07286-candle}} & \multicolumn{1}{c|}{CANDLE} & 73.8 & 64.7 & 77.6 & 59.4 & \multicolumn{1}{c|}{71.3} & 69.4 \\
\rowcolor{lightblue}
\multicolumn{1}{l|}{\textsc{Imagine}-GPT-2-L} & \multicolumn{1}{c|}{Synthetic VQA} & 61.5 & 63.9 & 68.9 & 53.0 & \multicolumn{1}{c|}{55.2} & 58.5 \\
\rowcolor{lightblue}
\multicolumn{1}{l|}{\textsc{Imagine}-RoBERTa-L} & \multicolumn{1}{c|}{Synthetic VQA} & 74.7 & 67.5 & 72.3 & 64.3 & \multicolumn{1}{c|}{61.2} & 68.0 \\
\rowcolor{lightblue}
\multicolumn{1}{l|}{\textsc{Imagine}-DeBERTa-v3-L} & \multicolumn{1}{c|}{Synthetic VQA} & \textbf{82.2} & \textbf{74.0} & \textbf{80.7} & {\ul 66.3} & \multicolumn{1}{c|}{76.7} & \textbf{76.0} \\ \midrule
\multicolumn{1}{l|}{Human} & \multicolumn{1}{c|}{-} & 91.4 & 88.9 & 94.9 & 86.9 & \multicolumn{1}{c|}{94.1} & 91.2 \\ \bottomrule
\end{tabular}%
}
\caption{Zero-shot evaluation results on commonsense reasoning tasks (Accuracy \%). \textbf{Bold} and 
{\ul Underline} indicate the best and second-best results, respectively. AT, CN, WD, WN, and AbsAT refer to ATOMIC, ConcetNet, WikiData, WordNet, and AbstractATOMIC. The full comparison is presented in Table \ref{tab:app_csqas} (Appendix). The results are from each reference.} 
\label{tab:csqas}
\end{table*}


\begin{table}[ht!]
\centering
\begin{adjustbox}{width=\linewidth}
\begin{tabular}{@{}lcccccc@{}}
\toprule
\multicolumn{1}{l|}{Method} & $\alpha$NLI & CSQA & PIQA & SIQA & \multicolumn{1}{c|}{WG} & Avg.  \\ \midrule
\multicolumn{1}{l|}{GPT-3.5}  & 61.8 & 68.9 & 67.8 & {\ul 68.0} & \multicolumn{1}{c|}{60.7} & 65.4  \\
\multicolumn{1}{l|}{ChatGPT} & 73.2 & \textbf{75.7} & {\ul 81.7} & \textbf{69.7} & \multicolumn{1}{c|}{64.1} & {\ul 72.9}  \\
\multicolumn{1}{l|}{GPT-4}  & {\ul 75.0} & 43.0 & 73.0 & 57.0 & \multicolumn{1}{c|}{\textbf{77.0}} & 65.0  \\
\multicolumn{1}{l|}{LLaMA2$_{\text{13B}}$} & 55.9 & 67.3 & 80.2 & 50.3 & \multicolumn{1}{c|}{72.8} & 65.3  \\
\multicolumn{1}{l|}{Mistral$_{\text{7B}}$} & 51.0 & 59.6 & \textbf{83.0} & 42.9 & \multicolumn{1}{c|}{75.3} & 62.4 \\
\rowcolor{lightblue}
\multicolumn{1}{l|}{\textsc{Imagine}}  & \textbf{82.2} & {\ul 74.0} & 80.7 & 66.3 & \multicolumn{1}{c|}{{\ul 76.7}} & \textbf{76.0} \\ \midrule
\multicolumn{1}{l|}{Human} & 91.4 & 88.9 & 94.9 & 86.9 & \multicolumn{1}{c|}{94.1} & 91.2 \\ \bottomrule
\end{tabular}%
\end{adjustbox}
\caption{Zero-shot evaluation results of LLMs on commonsense reasoning tasks (Accuracy \%). \textbf{Bold} and {\ul Underline} indicate the best and second-best results, respectively. Results are taken from \citet{DBLP:journals/corr/abs-2401-07286-candle}, and \textsc{Imagine} represents the results on DeBERTa-v3-L.}
\label{tab:csqas_llm}
\end{table}

\begin{table}[t!]
\centering
\resizebox{\columnwidth}{!}{%
\begin{tabular}{@{}l|cccc@{}}
\toprule
Method & QASC & SciQ & ARC-E & ARC-C \\ \midrule
SMLM$^\ast$ & 26.6 & - & 33.4 & 28.4 \\
CAR-RoBERTa-L & 56.7 & 60.7 & 57.0 & 36.5 \\
CAR-DeBERTa-v3-L & {\ul 70.0} & {\ul 76.9} & 75.3 & 53.2 \\ 
OPT$_{\text{30B}}$$^\ast$ & 39.7 & 72.7 & 58.2 & 34.8 \\
FLAN$_{\text{137B}}$$^\ast$ & - & - & \textbf{79.5} & \textbf{61.7} \\ \midrule
Z-LaVI (RoBERTa-L)$^\ast$ & 27.2 & 51.3 & 51.8 & 33.4 \\
Z-LaVI (BART-L)$^\ast$ & 27.3 & 51.0 & 56.1 & 36.5 \\
Z-LaVI (OPT$_{\text{30B}}$)$^\ast$ & 42.1 & 74.0 & 59.5 & 34.1 \\ 
\rowcolor{lightblue}
\textsc{Imagine}-GPT-2-L & 46.5 & 58.4 & 55.1 & 35.1 \\
\rowcolor{lightblue}
\textsc{Imagine}-RoBERTa-L & 57.1 & 63.7 & 57.9 & 39.1 \\
\rowcolor{lightblue}
\textsc{Imagine}-DeBERTa-v3-L & \textbf{72.4} & \textbf{78.9} & {\ul 76.0} & {\ul 56.2} \\ \bottomrule
\end{tabular}%
}
\caption{Zero-shot evaluation results on four science question-answering tasks (Accuracy \%). \textbf{Bold} and {\ul Underline} indicate the best and second-best results, respectively. Results ($^\ast$) are taken from references \citep{DBLP:conf/emnlp/BanerjeeB20-smlm, DBLP:conf/emnlp/YangYZWYC22-zlavi,  DBLP:conf/iclr/WeiBZGYLDDL22-flan}}
\label{tab:sciences}
\end{table}

\section{Experiments}
In this section, we demonstrate the effectiveness of \textsc{Imagine}. Specifically, we conduct extensive experiments and analysis to answer the following research questions:
\begin{itemize}
    \item [\textbf{Q1}] \textbf{(Generalizability)} Does \textsc{Imagine} offer better zero-shot performance across a broad range of reasoning benchmarks? (\S\ref{subsec: main})
    
    \item [\textbf{Q2}] \textbf{(Multimodality)} Does \textsc{Imagine} effectively integrate visual signals (imagination) with textual knowledge?  (\S\ref{subsec: insight}, \S\ref{subsec: insight2})
    
    \item [\textbf{Q3}] \textbf{(Effectiveness)} How effective are the components of \textsc{Imagine} in zero-shot commonsense reasoning? (\S\ref{subsec: trans})
    
\end{itemize}

\subsection{Experimental Setup}
\paragraph{Dataset.}
Following the previous works on zero-shot reasoning \citep{DBLP:conf/aaai/MaIFBNO21-ma, DBLP:conf/emnlp/YangYZWYC22-zlavi}, we evaluate our framework on commonsense reasoning tasks and science QA tasks to assess its generalizability \footnote{Evaluation results on NLU tasks are in Appendix \ref{app: nlu_tasks}.}.
Specifically, we evaluate each baseline on the five reasoning benchmarks, including Abductive NLI ($\alpha$NLI; \citealp{DBLP:conf/iclr/BhagavatulaBMSH20-anli}), CommonsenseQA (CSQA; \citealp{DBLP:conf/naacl/TalmorHLB19-csqa}), PhysicalIQA (PIQA; \citealp{DBLP:conf/aaai/BiskZLGC20-piqa}), SocialIQA (SIQA; \citealp{DBLP:conf/emnlp/SapRCBC19-socialiqa}), and Winogrande (WG; \citealp{DBLP:conf/aaai/SakaguchiBBC20-winogrande}). These datasets vary significantly in format (e.g., natural language inference, QA, pronoun resolution) and required knowledge (e.g., social and physical knowledge for SIQA and PIQA, respectively), enabling a comprehensive evaluation of a wide spectrum of reasoning capabilities. For science QA tasks, we assess each baseline on the four benchmarks, including QA via Sentence Composition (QASC; \citealp{DBLP:conf/aaai/KhotCGJS20-qasc}), Science Questions (SciQ; \citealp{DBLP:conf/aclnut/WelblLG17-sciq}), and the AI2 Reasoning Challenge (ARC-Easy, ARC-Challenge; \citealp{DBLP:journals/corr/abs-1803-05457-arc}). Given that science QA datasets often contain various types of reporting bias, such as color and shape biases, we selected these datasets to verify the efficacy of \textsc{Imagine} in mitigating reporting bias.

\paragraph{Baselines.}
We mainly compare \textsc{Imagine} with the following zero-shot commonsense reasoning frameworks:  MR \cite{DBLP:conf/aaai/MaIFBNO21-ma}, SMLM \cite{DBLP:conf/emnlp/BanerjeeB20-smlm}, Zero-shot Fusion \cite{DBLP:conf/naacl/KimKKAHY22}, CAR \cite{DBLP:conf/emnlp/WangF0XLSB23-car}, and the state-of-the-art framework, CANDLE \cite{DBLP:journals/corr/abs-2401-07286-candle}. To confirm the efficacy of training with machine imagination in \textsc{Imagine}, we also compare it with Z-LaVI \cite{DBLP:conf/emnlp/YangYZWYC22-zlavi}, which leverages machine imagination but does not include the training process. Beyond the reasoning framework based on KBs, we evaluate the recent LLMs, which include LLaMA2$_{\text{13B}}$ \cite{DBLP:journals/corr/abs-2307-092880-llama2}, Mistral$_{\text{7B}}$ (v0.1) \citep{DBLP:journals/corr/abs-2310-06825-mistral}, OPT$_{\text{30B}}$ \cite{DBLP:journals/corr/abs-2205-01068-opt}, FLAN$_{\text{137B}}$ \cite{DBLP:conf/iclr/WeiBZGYLDDL22-flan}, and the GPT families (i.e., GPT-3.5, ChatGPT (\texttt{gpt-3.5-turbo)}, GPT-4). 

\paragraph{Backbones.}
To verify the general applicability of \textsc{Imagine}, we apply our method to the both encoder and decoder models. Specifically, following the previous works, we utilize RoBERTa-Large \cite{DBLP:journals/corr/abs-1907-11692-roberta} and DeBERTa-v3-Large \cite{DBLP:conf/iclr/HeGC23-debertav3}. Each model has 362M and 443M parameters, respectively. As for the decoder model, we use GPT-2-Large that involves 792M parameters. Implementation details are in Appendix \ref{app: details}.

\begin{figure*}[t!]
\centering
  \includegraphics[width=\linewidth]{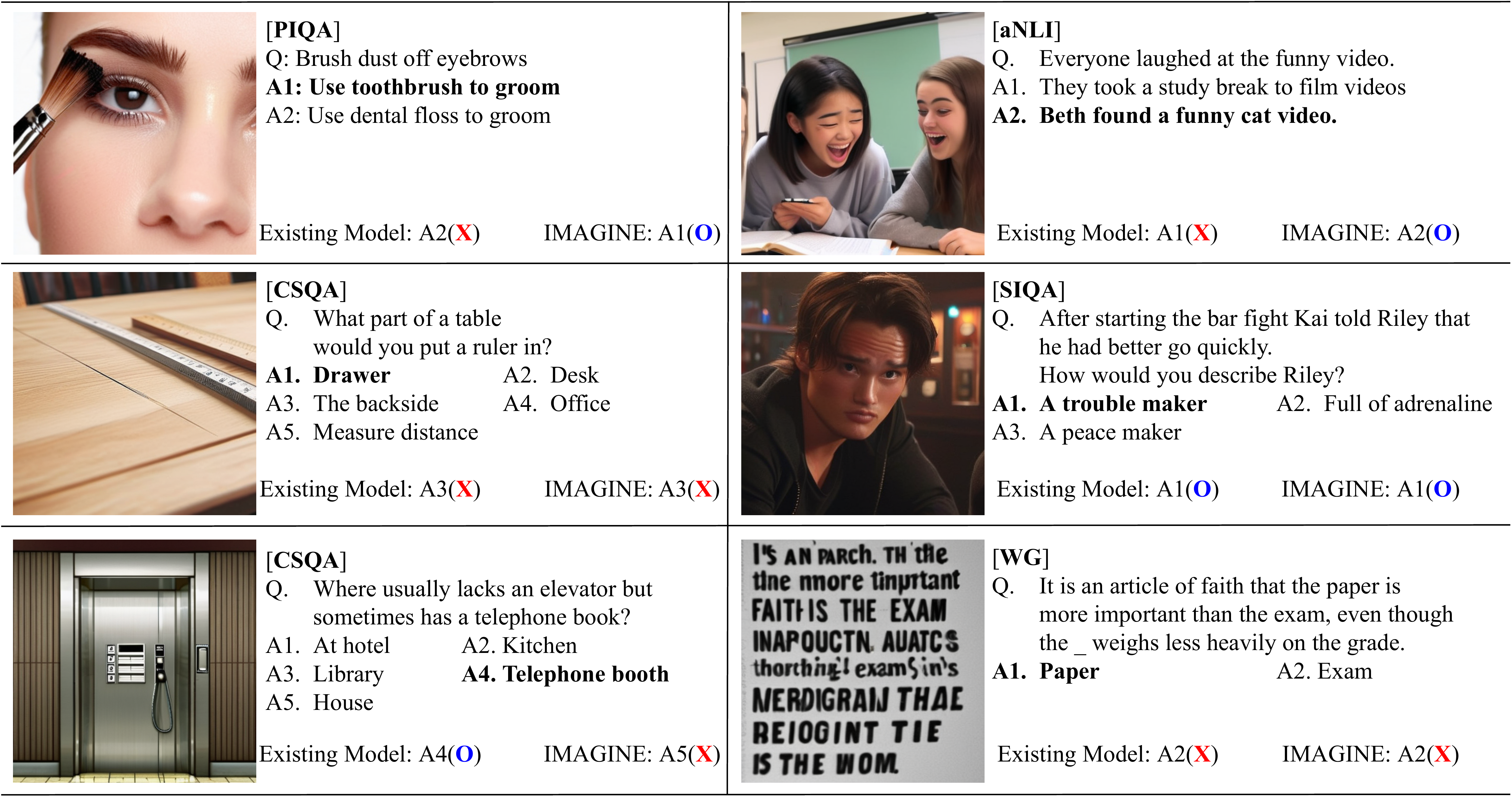}
    \caption{Comparison of model predictions and the correctness from \textsc{Imagine} and the existing model \citep{DBLP:conf/emnlp/WangF0XLSB23-car} on five commonsense reasoning tasks.}
    \label{fig:qual}
\end{figure*}

\subsection{Main Results}
\label{subsec: main}

Tables \ref{tab:csqas}, \ref{tab:csqas_llm}, and \ref{tab:sciences} show the results for the commonsense reasoning tasks and the science question-answering tasks. Models based on \textsc{Imagine} reveal either superior or competitive performance on overall reasoning tasks. This demonstrates the effectiveness of \textsc{Imagine} and highlights the benefit of leveraging machine imagination for reasoning. 

In particular, compared to zero-shot commonsense reasoning frameworks in commonsense reasoning tasks (Table  \ref{tab:csqas}), \textsc{Imagine}-DeBERTa-v3-L model surpasses the previous state-of-the-art by 0.9\%p on average, and specifically by 4.1\%p on the CSQA. This suggests that Synthetic VQA significantly enhances generalization performance in zero-shot commonsense reasoning. Comparison results with LLMs (Table \ref{tab:csqas_llm}) also shows that \textsc{Imagine} outperforms recent LLMs, including ChatGPT and GPT-4 \citep{DBLP:journals/corr/abs-2303-08774-gpt4}. This result suggests the superior efficiency and effectiveness of \textsc{Imagine}'s multimodal approach.

\textsc{Imagine} also proves effective for science QA tasks (Table \ref{tab:sciences}). Compared to the models with KBs and larger models, \textsc{Imagine} presents better or competitive reasoning performance. These results confirm the effectiveness of the machine imagination capabilities on science-related contexts. We also highlight the comparison results with Z-LaVI \cite{DBLP:conf/emnlp/YangYZWYC22-zlavi} that leverages imagination similar to ours. \textsc{Imagine} outperforms this method by a significant margin (18.5\%p on average), underscoring the importance of the pre-training phase in effectively utilizing machine imagination.

\begin{table}[t!]
\small
\centering
\begin{tabular}{@{}l|ccccc@{}}
\toprule
Imagine & aNLI & CSQA & PIQA & SIQA & WG \\ \midrule
Helpful (\%) & 30.2 & 41.1 & 26.9 & 36.2 & 11.8 \\
Harmful (\%) & 8.0 & 5.6 & 8.1 & 9.2 & 2.8 \\ \bottomrule
\end{tabular}%
\caption{Evaluation of reliance on machine-generated images using \textsc{Imagine}-DeBERTa-v3-L}
\label{tab: is_helpful_1}
\end{table}

\begin{table}[t!]
\small
\centering
\begin{tabular}{@{}l|ccccc@{}}
\toprule
Imagine & aNLI & CSQA & PIQA & SIQA & WG \\ \midrule
Helpful (\%) & 2.5 & 4.7 & 3.4 & 2.6 & 0.6 \\
Harmful (\%) & 1.7 & 3.7 & 2.7 & 1.3 & 0.5 \\ \bottomrule
\end{tabular}%
\caption{Evaluation of reliance on machine-generated images using \textsc{Imagine}-RoBERTa-L}
\label{tab: is_helpful_2}
\end{table}

\subsection{Impact of Imagination on Model Inference}\label{subsec: insight}
We analyze the inference results from the text-based model, CAR \citep{DBLP:conf/emnlp/WangF0XLSB23-car}, and \textsc{Imagine} to confirm the impact of machine imagination on the model inference. The results are shown in Figure \ref{fig:qual}. We draw three major findings regarding the impact of imagination: (i) When the text contains limited commonsense knowledge, imagination indeed helps the model to correctly infer the answer (First row in the Figure), i.e., positive impact on predictions (ii) When the generated images only partially capture the context of the text query, imagination does not affect the inference results (Second row in the Figure). (iii) When images deviate from the real world, imagination can lead to incorrect inferences (Third row in the Figure). Specifically, we empirically observe that longer text queries often result in such cases. 

To further assess how often images negatively impact model inference, we calculate the ratio of helpful imagination (i.e., imagination leading to correct reasoning) to harmful imagination (i.e., imagination leading to incorrect reasoning) across different commonsense reasoning benchmarks (Table \ref{tab: is_helpful_1} and \ref{tab: is_helpful_2}). Our analysis shows that helpful imagination contributes more than harmful imagination, suggesting that imagination generally has a positive impact. However, we also observe that in certain cases, misaligned imagination can lead to reasoning errors.

These results suggest that incorporating a text-to-image model with better alignment capabilities could potentially mitigate the negative impacts of imagination. We provide more examples with the visualization of model attention in Appendix \ref{app: Attention}.

\begin{table}[t!]
\small
\centering
\resizebox{\columnwidth}{!}{%
\begin{tabular}{@{}l|ccccc|c@{}}
\toprule
KB       & $\alpha$NLI & CSQA & PIQA & SIQA & WQ   & Avg. \\ \midrule
Synthetic VQA  & {\ul 74.7} & {\ul 67.5} & {\ul 72.3} & {\ul 64.3} & \textbf{61.2} & \textbf{68.0} \\ \midrule
{\small w/o} VCR        & 71.7 & 65.7 & \textbf{72.3} & \textbf{65.7} & {\ul 60.3} & {\ul 67.1} \\
{\small w/o} AbsAT      & \textbf{75.6} & \textbf{67.5} & 71.7 & 56.2 & 58.8 & 66.0 \\
{\small w/o} VCR, AbsAT & 65.6 & 45.0 & 67.6 & 47.3 & 57.5 & 56.6 \\ \bottomrule
\end{tabular}
}
\caption{Ablation results on Synthetic VQA. \textbf{Bold} and {\ul underline} indicate the best and second-best results.}
\label{tab:kb_ablation}
\end{table} 

\subsection{Contributions of Synthetic VQA}\label{subsec: insight2}
To confirm the effectiveness of each component in Synthetic VQA, we evaluate the contribution of AbsAT and VCR. Table \ref{tab:kb_ablation} presents the results on commonsense reasoning tasks. The model trained only with AbsAT (i.e., w/o VCR) shows superior performance on datasets that contain longer sequences and require complex knowledge (e.g., PIQA, SIQA). In contrast, the model trained only with VCR (i.e., w/o AbsAT) shows its strength on the dataset that contain simpler questions ($\alpha$NLI, CSQA) which allows the better use of visual information. When combining these two components, the Synthetic VQA results in well-generalized reasoners across diverse reasoning tasks, demonstrating the complementary effect of each component.

\begin{table}[t!]
\footnotesize
\centering
\resizebox{\columnwidth}{!}{%
\begin{tabular}{@{}cc|ccccc|c@{}}
\toprule
LM & ITM & $\alpha$NLI & CSQA & PIQA & SIQA & WG & Avg. \\ \midrule
\checkmark & \checkmark & \textbf{74.7} & \textbf{67.5} & \textbf{72.3} & \textbf{64.3} & \textbf{61.2} & \textbf{68.0} \\
\checkmark & - & 74.3 & 65.2 & 71.9 & 62.3 & 60.5 & 66.8 \\
- & \checkmark & 71.7 & 62.0 & 68.8 & 60.0 & 59.6 & 64.4 \\
- & - & 65.6 & 45.0 & 67.6 & 47.3 & 57.5 & 56.6 \\ \bottomrule
\end{tabular}%
}
\caption{Ablation results on pre-training objective of \textsc{Imagine}. We use a RoBERTa-L as a backbone.}
\label{tab:objective_ablation}
\end{table}

\begin{table}[t!]
\footnotesize
\centering
\resizebox{\columnwidth}{!}{%
\begin{tabular}{@{}l|ccccc|c@{}}
\toprule
{Inference} & $\alpha$NLI & CSQA & PIQA & SIQA & WG & Avg. \\ \midrule
Ensemble & \textbf{74.7} & \textbf{67.5} & \textbf{72.3} & \textbf{64.3} & \textbf{61.2} & \textbf{68.0} \\
LM & 74.1 & 66.9 & 71.8 & 63.8 & 61.1 & 67.1 \\
ITM & 71.7 & 63.1 & 68.3 & 59.8 & 59.4 & 64.0 \\ \bottomrule
\end{tabular}%
}
\caption{Results of the different inference strategy (LM, ITM). These strategies are evaluated on RoBERTa-L.}
\label{tab:scoring}
\end{table}

\subsection{Component Analysis on \textsc{Imagine}}
\label{subsec: trans}
\paragraph{Ablation on Training Objectives.}
\textsc{Imagine} employs two objectives (i.e., LM, ITM) to learn commonsense knowledge from different modalities. We perform ablations on these objectives to verify their contributions in enhancing zero-shot reasoning capabilities. Table \ref{tab:objective_ablation} shows the ablation results. Notably, omitting the LM objective leads to a significant drop in performance, underscoring the crucial role of language understanding in commonsense reasoning. Furthermore, while ITM alone does not significantly impact reasoning effectiveness, combining ITM with LM results in improved reasoning performance. These findings suggest that integrating visual information in model optimization leads to better reasoning in commonsense situations.

\paragraph{Effect of Ensemble Inference.} 
\textsc{Imagine} performs reasoning by ensembling LM and ITM scores. To investigate the contributions in scores obtained from these two different modalities, we evaluate each score independently. The results are presented in Table \ref{tab:scoring}. We observe the lowest performance when evaluating only the ITM scores. However, ensembling LM scores with the ITM results in significant performance improvement across all tasks, even though the scores derived from images are much lower than those from text. This indicates that integrating machine-generated images can complement and enhance language-based reasoning abilities. More analysis on ensemble methods are in Appendix \ref{app:ensemble}.

\paragraph{Impact of Adapter.}
\textsc{Imagine} utilizes parallel adapters \cite{DBLP:conf/iclr/HeZMBN22-parallel_adapter} to alleviate the conflicts between the two objectives (i.e., LM, ITM) during the pre-training. In this study, we examine whether separating parameters through adapters for distinct modality objectives is truly effective. Table \ref{tab:adapter} presents the ablation results on adapters. We observe a significant decline in reasoning performance when adapters are removed. This suggests that direct training of PLMs with images adversely affects the acquisition of textual knowledge. One plausible explanation for this phenomenon is possibly related to catastrophic forgetting \citep{kirkpatrick2017overcoming}, where the model loses previously acquired knowledge (i.e., textual knowledge inherent in PLMs). This highlights the effectiveness of adapters in maintaining the model’s linguistic understanding when it learns from new modalities.

\subsection{Comparison with VL models}
\label{subsec: vlmodels}
We include the state-of-the-art language models as baselines (e.g., GPT-4, LLaMA2), as our focus is on enhancing language-based reasoning ability using visual signals. Nevertheless, we also provide results from recent powerful vision-language (VL) models (LLaVA-1.5 \citep{DBLP:conf/nips/LiuLWL23a}, InstructBLIP with Vicuna-7B \citep{DBLP:conf/nips/Dai0LTZW0FH23-instructblip}) by feeding the generated images from our framework. The results in Table \ref{tab:vlmodels} indicate that these VL models struggle to reason accurately about commonsense questions. We suspect that this issue arises from VL models’ tendency to focus on the image scene more than on textual inputs, as they are primarily trained to answer questions based on the entire image scene.
The datasets we experiment with prioritize linguistic ability over vision-language grounding and require reasoning rooted in commonsense knowledge. As a result, VL models that are more focused on visual understanding may underperform in zero-shot commonsense reasoning tasks, where strong linguistic reasoning is crucial.

\begin{table}[t!]
\small
\centering
\resizebox{\columnwidth}{!}{%
\begin{tabular}{@{}l|ccccc|c@{}}
\toprule
 Model & $\alpha$NLI & CSQA & PIQA & SIQA & WG & Avg. \\ \midrule
{Parallel Adapter} & \textbf{74.7} & \textbf{67.5} & \textbf{72.3} & \textbf{64.3} & \textbf{61.2} & \textbf{68.0} \\ 
{Full} & 73.0 & 65.4 & 71.1 & 61.5 & 61.2 & 66.4 \\
\bottomrule
\end{tabular}%
}
\caption{Evaluation results of \textsc{Imagine} with full fine-tuning (Full) and adapter tuning (Adapter).}
\label{tab:adapter}
\end{table}

\begin{table}[t!]
\small
\centering
\resizebox{\columnwidth}{!}{%
\begin{tabular}{@{}l|ccccc|c@{}}
\toprule
Method & $\alpha$NLI & CSQA & PIQA & SIQA & WG & Avg. \\ \midrule
LLaVA$_{\text{7B}}$ & 55.2 & 29.4 & 64.2 & 34.8 & 54.5 & 47.6 \\
InstructBLIP$_{\text{7B}}$ & 54.8 & 40.5 & 66.0 & 42.1 & 59.6 & 52.6 \\
CANDLE & 81.2 & 69.9 & 80.3 & 65.9 & \textbf{78.3} & 75.1 \\
\textsc{Imagine} & \textbf{82.2} & \textbf{74.0} & \textbf{80.7} & \textbf{66.3} & 76.7 & \textbf{76.0} \\ \bottomrule
\end{tabular}%
}
\caption{Zero-shot evaluation results of VL models and \textsc{Imagine} on commonsense reasoning tasks (Accuracy \%). The backbone of \textsc{Imagine} and CANDLE is DeBERTa-v3-L.}
\label{tab:vlmodels}
\end{table}

\section{Conclusion}

In this paper, we have proposed \textsc{Imagine}, a novel zero-shot commonsense reasoning framework that leverages visual signals to mitigate reporting bias in textual inputs.
To steer \textsc{Imagine} in effectively utilizing visual information, we have created a large-scale Synthetic VQA dataset and optimized the model to use both textual and visual information. 
Our extensive experiments have shown that \textsc{Imagine} establishes new state-of-the-art results on zero-shot commonsense reasoning tasks compared to strong baselines (including large language models), demonstrating the efficacy of machine imagination. 
Moreover, the in-depth analysis clearly supports the strength of the proposed method by showing that the model tends to utilize visual information beneficially. 



\section*{Limitations}
We have demonstrated the efficacy of the machine imagination to improve zero-shot commonsense reasoning ability. However, we still have the following limitations:

\paragraph{Additional Computations}
While machine imagination leads to performance improvement in PLMs, it necessitates additional computations for generating and processing visual signals. This limitation can be addressed by retrieving relevant images instead of synthesizing new ones, as demonstrated in previous work \cite{DBLP:conf/emnlp/YangYZWYC22-zlavi}. We consider this approach a promising avenue for future research.

\paragraph{Exploration of \textsc{Imagine} on LLMs}

In this work, we apply \textsc{Imagine} to only intermediate-size models (300M to 790M), as one of our objectives is to see if the smaller models with machine imagination outperform LLMs on a broad range of commonsense reasoning tasks. This objective motivates us to apply our method to language models with less than 1B parameters. Additionally, from a practical perspective, the proposed method involves a pre-training phase to teach the joint use of multi-modal data. This process requires substantial computational costs to train larger models. However, we believe that \textsc{Imagine} can be effectively combined with LLMs, given that the reporting bias is an inherent issue in the pre-training corpus and not the models themselves. We plan to explore the scaling of machine imagination in our future research.

\section*{Acknowledgments}
This work was supported by the National Research Foundation of Korea (NRF) grant funded by the Korea government (MSIT) (No.RS-2024-00415812 and No.2021R1A2C3010430) and Institute of Information \& communications Technology Planning \& Evaluation (IITP) grant funded by the Korea government (MSIT) (No.RS-2024-00439328, Karma: Towards Knowledge Augmentation for Complex Reasoning (SW Starlab), No.RS-2024-00457882, AI Research Hub Project, and No.RS2019-II190079, Artificial Intelligence Graduate School Program (Korea University)).   

\bibliography{custom}

\clearpage
\appendix

\begin{center}
\textbf{Appendix}    
\end{center}

\section{Synthetic VQA dataset}
\label{app:data}

\begin{table}[ht!]
\centering
\resizebox{\columnwidth}{!}{%
\begin{tabular}{@{}l|cc|c@{}}
\toprule
 & Train & Dev & Total \\ \midrule
\# Images generated from AbsAT & 18,838 & 1,695 & 20,533 \\
\# QA pairs from AbsAT & 486,778 & 46,238 & 533,016 \\ \midrule
\# Images from VCR & 80,418 & 9,929 & 90,347 \\
\# QA pairs from VCR & 212,923 & 26,534 & 239,457 \\ \midrule
\# Total Images & 99,256 & 11,624 & 110,880 \\
\# Total QA pairs & 699,701 & 72,772 & 772,473 \\ \bottomrule
\end{tabular}%
}
\caption{Statistic of Synthetic VQA dataset.}
\label{tab:statistics}
\end{table}

We construct a Synthetic VQA dataset using AbstractATOMIC and VCR. First, we generate images using the questions from AbstractATOMIC. Since AbstractATOMIC consists only of text, we need to create images based on these questions. In this process, we standardize all the person names in the questions to ``Person'' and remove duplicate questions, resulting in approximately 20K images.
To include more realistic images and commonsense questions corresponding to those images, we extract question-answer pairs from VCR images. However, most of these questions are directly related to the images, making it difficult to answer without them, which poses a challenge for LM-based training. To address this, we replace the person indices in the questions with gender-neutral names and generate captions for the images to use as prefixes for the questions. In addition, each QA pair from VCR has four answer candidates, while each pair from AbstractATOMIC has three candidates. To combine them, we match the number of answer choices by randomly discarding one distractor from VCR. The statistic of our dataset is provided in Table \ref{tab:statistics}.

\section{Implementation Details}
\label{app: details}
To construct the VQA pairs, we primarily use DALL-E 3-XL \citep{betker2023improving-dalle3}, a powerful image synthesis model. For generating images in the Synthetic VQA dataset, we first remove overly specific information, such as personal names, from the questions. Then, we generate images with a resolution of $384\times384$ using 50 inference steps. During the evaluation, we generate $512\times512$ images for each task based on the questions, maintaining the same number of inference steps.
We use the CLIP-Large \citep{DBLP:conf/icml/RadfordKHRGASAM21-clip} model to extract image features. Following prior work, we use two powerful PLMs as the backbone. We add Parallel Adapter \citep{DBLP:conf/iclr/HeZMBN22-parallel_adapter} with a reduction factor of 16 to each model and freeze all parameters except for the adapters. We follow the training settings of \citet{DBLP:conf/aaai/MaIFBNO21-ma} and \citet{DBLP:conf/emnlp/WangF0XLSB23-car} to train Transformer decoder-based and encoder-based model for the in-depth comparison. We report our results derived from the ensemble score using the optimal ensemble weight for each task.
All experiments are conducted using four NVIDIA A5000 GPUs. More details are presented in Table \ref{tab:implementation_details}.

\begin{table}[!t]
\centering
\resizebox{\columnwidth}{!}{%
\begin{small}
\begin{tabular}{@{}l|ccc@{}}
\toprule
\textsc{Imagine} & GPT-2-L & RoBERTa-L & DeBERTa-v3-L \\ \midrule
Image Encoder & \multicolumn{3}{c}{CLIP-ViT-L/14} \\
\# Params. & 792M + 428M & 362M + 428M & 443M + 428M \\
\# Trainable Params. & 7.9M & 8.4M & 8.4M \\
Training Time & 70h & 30h & 80h \\  
Batch Size & \multicolumn{3}{c}{8, 16, \textbf{32}, 64} \\
Learning Rate & \multicolumn{3}{c}{7e-6, \textbf{1e-5}, 3e-5} \\
Epoch & \multicolumn{3}{c}{2} \\ \bottomrule

\end{tabular}
\end{small}
}
\caption{Detailed training settings for \textsc{Imagine}. \textbf{Bold} indicates the chosen hyperparameter.}
\label{tab:implementation_details}
\end{table}

\section{Ensemble Methods}
\label{app:ensemble}
To verify the effectiveness of our framework's multimodality approach, we train two unimodal models using different seeds on the Synthetic VQA dataset, utilizing only the text. We then ensemble the scores obtained from these two models. The results are presented in Table \ref{tab:score ensemble}. While ensembling scores from single modalities (LM+LM) provides performance benefits, ensembling scores from two different modalities (LM+ITM), as done in \textsc{Imagine}, proves to be the most effective. This demonstrates that the multimodality approach plays a crucial role in enhancing zero-shot reasoning performance.

\begin{table}[ht!]
\centering
\resizebox{\columnwidth}{!}{%
\begin{tabular}{@{}l|ccccc|c@{}}
\toprule
RoBERTa-Large & $\alpha$NLI & CSQA & PIQA & SIQA & WG & Avg. \\ \midrule
LM & 74.3 & 65.2 & 71.9 & 62.3 & 60.5 & 66.8 \\
LM+LM & 74.3 & 66.0 & 72.1 & 64.2 & 60.4 & 67.4 \\
LM+ITM (\textsc{Imagine}) & \textbf{74.7} & \textbf{67.5} & \textbf{72.3} & \textbf{64.3} & \textbf{61.2} & \textbf{68.0} \\ \bottomrule
\end{tabular}%
}
\caption{Results of two different ensemble methods.}
\label{tab:score ensemble}
\end{table}

We report the optimal ensemble weights used for our framework in Figure \ref{fig:ensemble_weight}. The larger the ensemble weight, the greater the influence of the image scores. Additionally, we draw a line indicating the average accuracy in each plot. From this, we can infer that the DeBERTa-v3-Large model utilizes image information more extensively than the RoBERTa-Large. When applying \textsc{Imagine} to DeBERTa-v3-Large, the performance improvement is greater than when using RoBERTa-Large, suggesting that visual information contributes positively to most reasoning tasks.

\section{Impact of Image Quality}
\label{app: c}
We aim to observe the changes in inference performance based on image quality by generating images of various qualities using three different methods. First, similar to our main experiment, we utilize the questions from the evaluation dataset to generate images with a resolution of $512\times512$ using both DALL-E 3-XL and the Latent Diffusion Model (LDM; \citealp{DBLP:conf/cvpr/RombachBLEO22-stable}), which has relatively lower image synthesis capabilities. Additionally, we generate images with a resolution of $384\times384$ using DALL-E 3-XL, following the same method used for creating the Synthetic VQA dataset.

\begin{figure}[t!]
\centering
  \includegraphics[width=\linewidth]{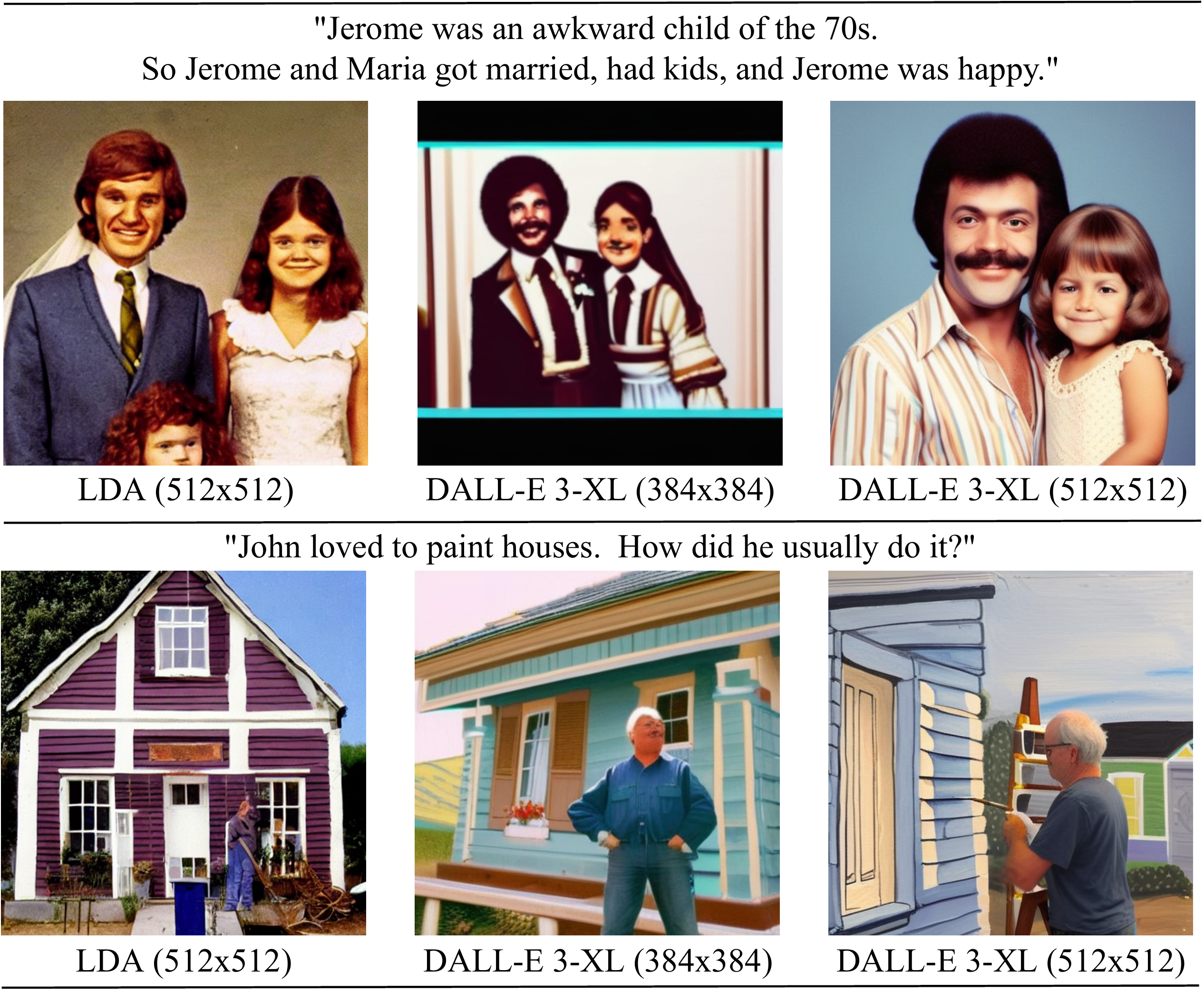}
    \caption{Comparison of generated images. The sentences are the queries used to generate the images.}
    \label{fig:img_quality}
\end{figure}

\begin{figure*}[ht!]
\centering
  \includegraphics[width=\linewidth]{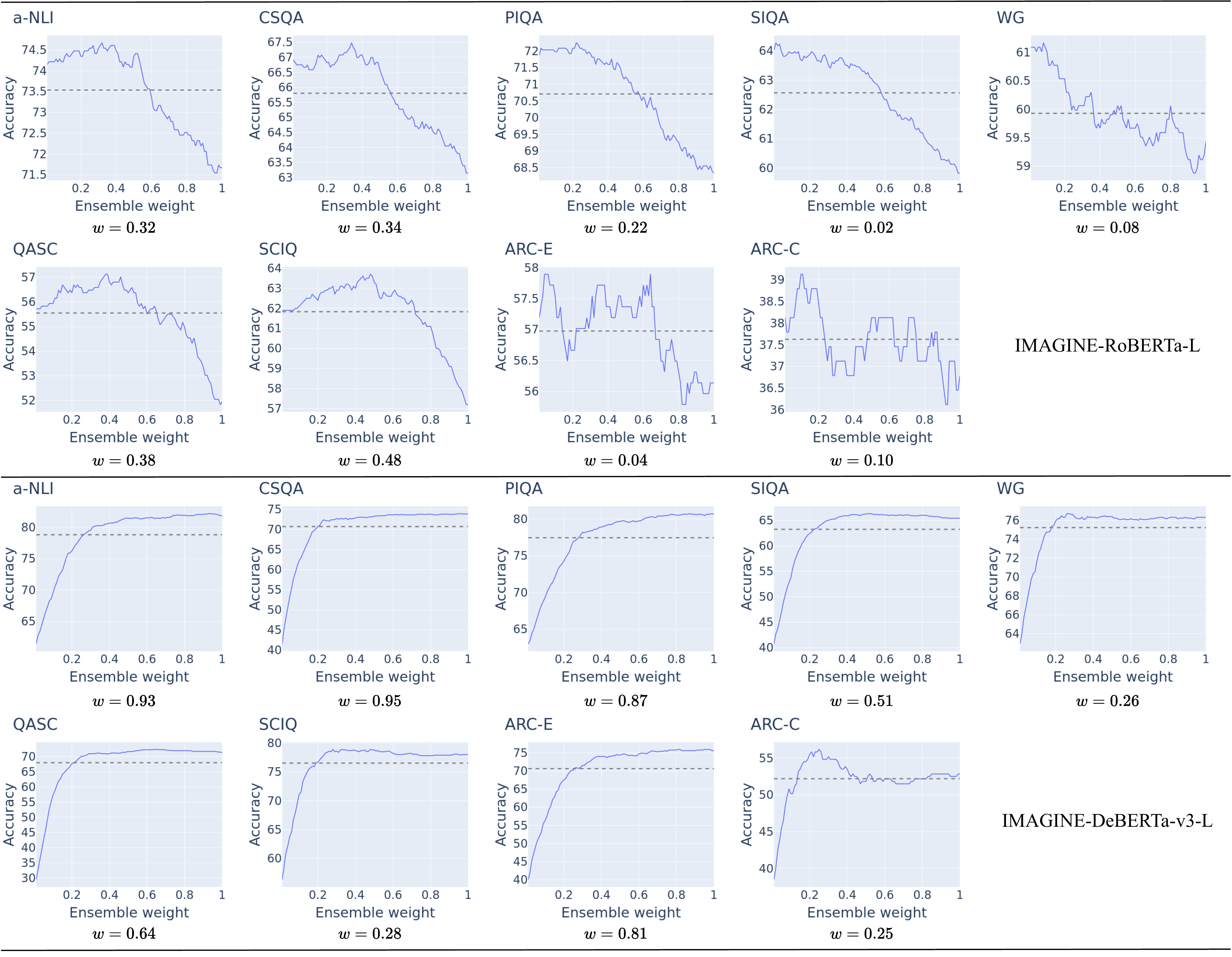}
    \caption{Model accuracy variation with different ensemble weights. The optimal $w$ for each task is shown below the plots. The line in the middle indicates the average accuracy.}
    \label{fig:ensemble_weight}
\end{figure*}

\begin{table}[ht!]
\centering
\resizebox{\columnwidth}{!}{%
\begin{tabular}{@{}l|ccccc|c@{}}
\toprule
\textsc{Imagine} & $\alpha$NLI & CSQA & PIQA & SIQA & WG & Avg. \\ \midrule
Text only & 73.2 & 66.3 & 71.3 & 64.5 & 60.3 & 67.1 \\
LDM ($512\times512$) & 73.2 & 66.3 & 71.9 & 64.3 & 60.6 & 67.3 \\
DALL-E 3 ($384\times384$) & 74.5 & 66.8 & 71.9 & 64.3 & 60.6 & 67.6 \\
DALL-E 3 ($512\times512$) & \textbf{74.7} & \textbf{67.5} & \textbf{72.3} & \textbf{64.3} & \textbf{61.2} & \textbf{68.0} \\ \bottomrule
\end{tabular}%
}
\caption{Results of using various image synthesis models for evaluation. The numbers in parentheses indicate the image resolution.}
\label{tab:img_quality}
\end{table}
The results in Table \ref{tab:img_quality} show that the \textsc{Imagine} with the LDM model performs the worst, indicating that utilizing a less effective image synthesis model can degrade overall performance. However, all models benefit from incorporating various resolutions of images. As seen in Figure \ref{fig:img_quality}, this is likely because the generated images, despite varying in quality, mostly maintain contextual relevance to the query sentences, thereby having a similar positive impact on the inference results.

\section{\textsc{Imagine} with Decoder-based Model}
\label{app:decoder}
We conducted experiments using GPT-2, a widely-used decoder-based generative language model, to verify the applicability to recent language models. We follow the settings of \citep{DBLP:conf/aaai/MaIFBNO21-ma} to train to model on synthetic datasets.
\begin{table}[ht!]
\centering
\resizebox{\columnwidth}{!}{%
\begin{tabular}{@{}l|ccccc|c@{}}
\toprule
 & $\alpha$NLI & CSQA & PIQA & SIQA & WG & Avg. \\ \midrule
GPT-2-L & 56.5 & 41.4 & 68.9 & 44.6 & 53.2 & 52.9 \\
GPT-2-L (MR) & 59.2 & 48.0 & 67.5 & 53.6 & 54.7 & 56.6 \\
CAR-GPT-2-L & 61.7 & 50.0 & 68.2 & 52.3 & 55.2 & 57.5 \\
\textsc{Imagine}-GPT-2-L & 61.5 & 53.9 & 68.9 & 53.0 & 55.2 & 58.5 \\
\bottomrule
\end{tabular}%
}
\caption{Zero-shot evaluation results with decoder-only generative model.}
\label{tab:decoder}
\end{table}

The results in Table \ref{tab:decoder} demonstrate that \textsc{Imagine} is effective not only for encoder-based models but also for decoder-based models. Based on these findings, we plan to address methodologies in future work that can effectively utilize images while preserving the rich language understanding capabilities of large language models.

\begin{table*}[ht!]
\small
\centering
\resizebox{\textwidth}{!}{%
\begin{tabular}{@{}l|cc|cccccccccc@{}}
\toprule
 & AOKVQA & VCR & Synthetic VQA & $\alpha$NLI & CSQA & PIQA & SIQA & WG & QASC & SciQ & ARC-E & ARC-C \\ \midrule
Relevance & 23.81 & 21.26 & 23.59 & 30.26 & 29.38 & 30.80 & 29.92 & 29.26 & 29.21 & 21.23 & 20.26 & 19.98 \\ \bottomrule
\end{tabular}%
}
\caption{Image-text relevance evaluation using CLIP-base model.}
\label{tab:relevance}
\end{table*}

\section{Validation of Synthetic Dataset Quality}
We evaluate the quality of the dataset by measuring the relevance between the question text and the machine-generated images. For this purpose, inspired by the CLIP scores \citep{DBLP:conf/emnlp/HesselHFBC21}, we measure the relevance score between images and text using the CLIP model. A higher relevance score between the two modalities indicates that the image effectively captures the content of the text. As shown in Figure \ref{fig:qual}, images that are highly relevant to the questions can help to reason about the question.

First, we measure the relevance of datasets containing two sets of real images (A-OKVQA, VCR) to establish a baseline. Then we compare these scores with those of the Synthetic VQA and the synthetic pairs of all evaluation datasets to determine the quality of the synthetic dataset. The results in Table \ref{tab:relevance} show that most datasets exhibit similar or even higher relevance scores compared to the datasets containing real images (A-OKVQA, VCR). In particular, for Synthetic VQA, we evaluate only the dataset extracted from AbstractATOMIC, which contains only machine-generated images, and found that it has relevance scores closest to those of the real-image datasets A-OKVQA and VCR. This demonstrates that our synthetic dataset has a quality comparable to that of the real VL dataset.

\section{Visualization of Image Attention}
\label{app: Attention}
\begin{figure*}[ht!]
\centering
  \includegraphics[width=\linewidth]{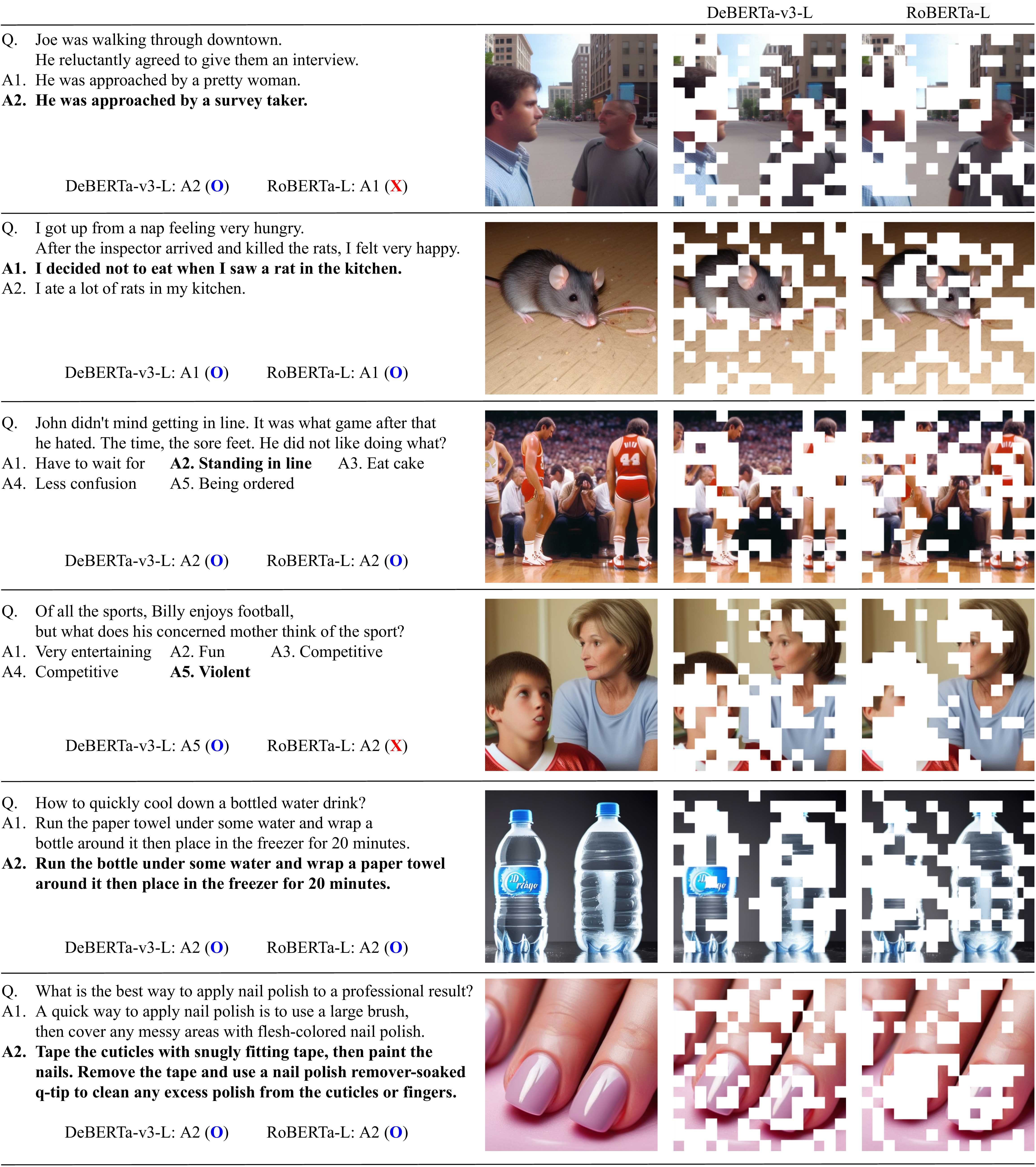}
    \caption{Randomly sampled examples from \textsc{Imagine} alongside the visualization of image attention from the Abductive NLI, CommonsenseQA, and PIQA validation sets.}
    \label{fig:ex1}
\end{figure*}

\begin{figure*}[ht!]
\centering
  \includegraphics[width=\linewidth]{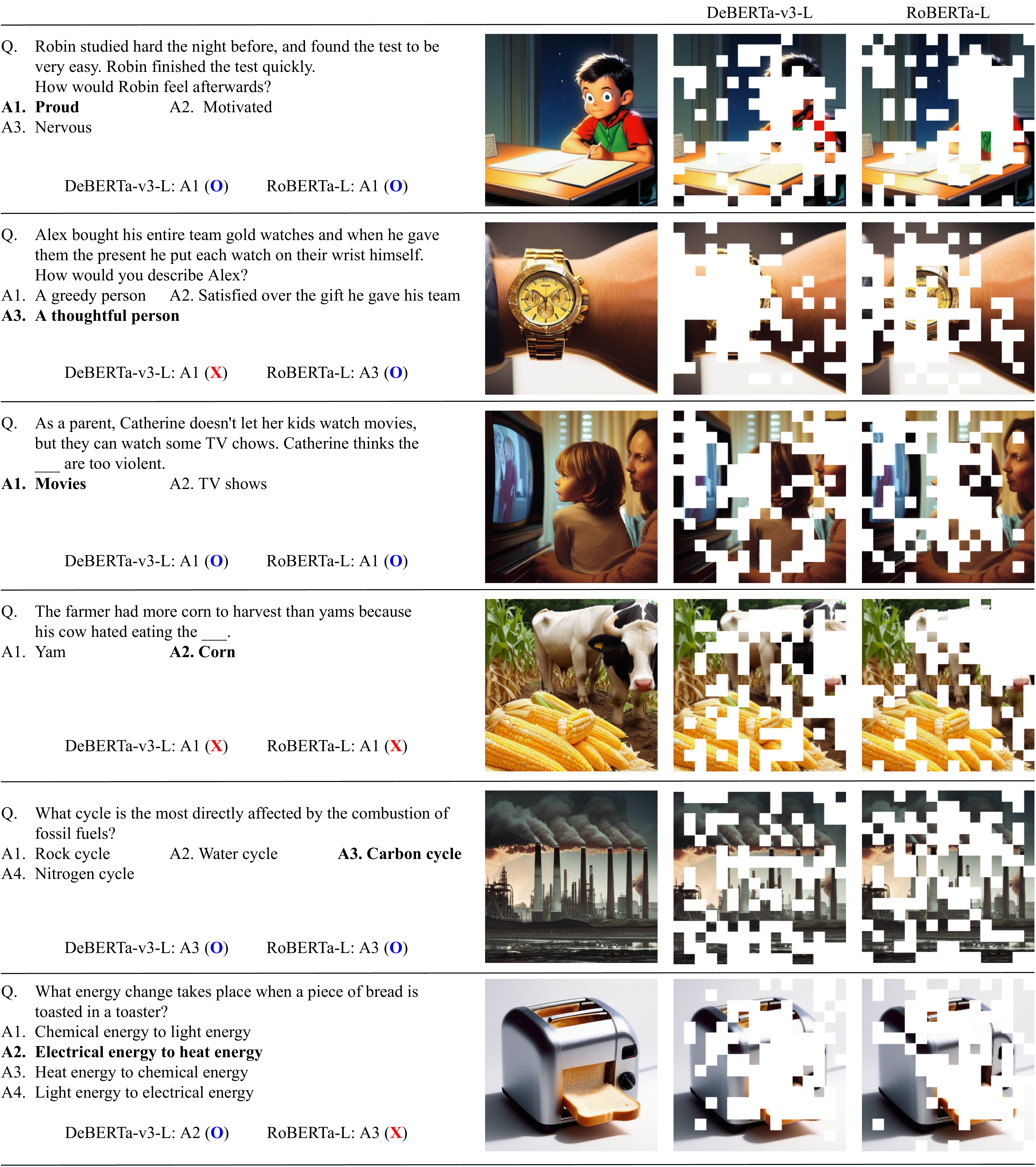}
    \caption{Randomly sampled examples from \textsc{Imagine} alongside the visualization of image attention from the SIQA, Winogrande, and ARC-easy validation sets.}
    \label{fig:ex2}
\end{figure*}

\begin{figure*}[ht!]
\centering
  \includegraphics[width=\linewidth]{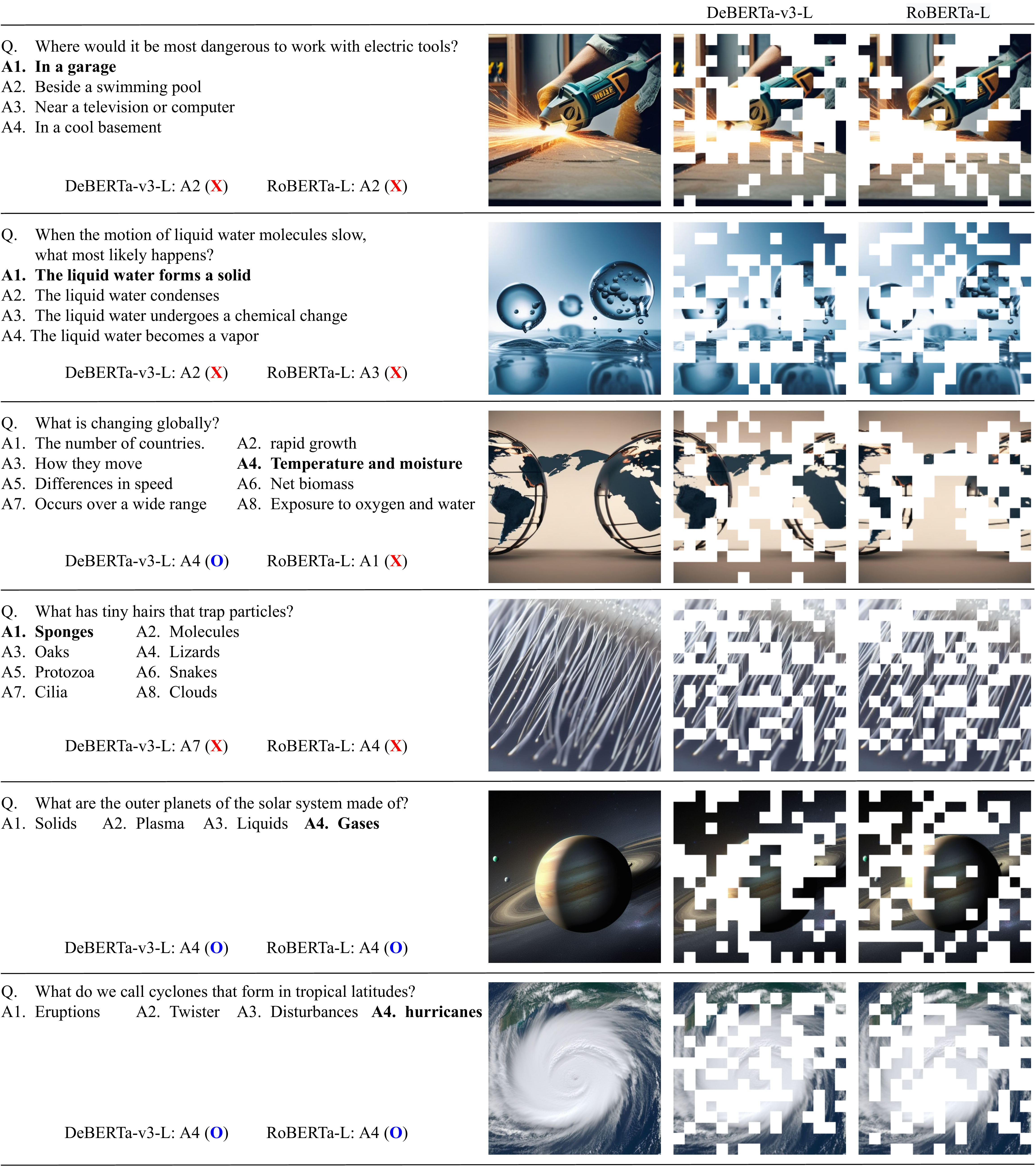}
    \caption{Randomly sampled examples from \textsc{Imagine} alongside the visualization of image attention from the ARC-challenge, QASC, and SciQ validation sets.}
    \label{fig:ex3}
\end{figure*}

We aim to visualize how the model utilizes specific parts of an image. The formula to compute contextualized visual features used for computing the ITM score calculation process is similar to the attention algorithm, allowing us to derive attention scores for each image patch. Based on these scores, we erase 100 image patches with the lowest scores to understand which parts the model focuses on. As shown in Figure \ref{fig:ex1}, \ref{fig:ex2}, and \ref{fig:ex3}, each model tends to assign relatively high attention scores to objects related to the question in most cases, rather than using the image patches randomly. This is notable because the model can effectively capture the relationship between text and images using adapters, despite training with much less data compared to existing visual-language modeling studies \citep{DBLP:conf/icml/0008LSH23-blip2, DBLP:journals/corr/abs-2304-10592-minigpt4}. In addition, we observe that the DeBERTa-v3-Large model tends to focus more frequently on the correct parts than the RoBERTa-Large model. Figure \ref{fig:ex1} shows these cases clearly. This aligns with the result that the \textsc{Imagine} is more effective with DeBERTa-v3-Large, suggesting that a model with high generalization performance is also useful for learning new modalities.


\section{Comparison of Inference Time}
Since \textsc{Imagine} utilizes both images and text after generating images, inference may take longer. Nevertheless, to demonstrate the effectiveness of our methodology, we analyze the model in terms of inference time.
The inference time and the number of parameters required to produce an answer vary depending on the setting, particularly the image quality. As shown in Appendix \ref{app: c}, when generating images using the LDM model and then inferring the answer with the \textsc{Imagine}-RoBERTa-L framework, the average time taken is 4.5 seconds (image generation: 4 seconds, image processing: 0.2 seconds, text processing: 0.3 seconds). The total number of parameters used is 1.7 billion (LDM: 1 billion, CLIP: 428 million, RoBERTa: 362 million). This model achieves higher performance with significantly fewer parameters compared to the 7 billion parameter large language models shown in Table 2. Although the 7 billion parameter models have an average inference speed of 2.1 seconds, we believe this is justified by the superior performance of our model.

Additionally, our largest setting (\textsc{Imagine}-DeBERTa-v3-Large framework containing DALL-E 3-XL) takes a total of 21.5 seconds to infer an answer and has 4.6 billion parameters. This model can achieve higher performance than large language models with over 30 billion parameters. This suggests that our framework is a more effective alternative to simply increasing model size.

\section{Versatility of \textsc{Imagine}}
\label{app: nlu_tasks}
To confirm the versatility of \textsc{Imagine}, we measure the performance of \textsc{Imagine} not only on zero-shot commonsense reasoning but also on several tasks from the GLUE dataset (SST-2, RTE) \citep{DBLP:conf/iclr/WangSMHLB19}. 

\begin{table}[ht!]
\small
\centering
\begin{tabular}{@{}lcc@{}}
\toprule
Method & SST-2 & RTE \\ \midrule
DeBERTa-v3-L & 49.1 & 50.5 \\
CAR-DeBERTa-v3-L & 56.2 & 52.3 \\
\textsc{Imagine}-DeBERTa-v3-L & 91.1 & 53.8 \\ \bottomrule
\end{tabular}%
\caption{Zero-shot evaluation resutls on natual language understanding tasks.}
\label{tab: nlu_tasks}
\end{table}

As shown in the Table \ref{tab: nlu_tasks}, the results indicate that the \textsc{Imagine}-DeBERTa-v3-L model achieves the highest performance across general tasks on average, suggesting that \textsc{Imagine} can indeed be a general approach to engage PLMs. Specifically, \textsc{Imagine} shows greater performance improvements in datasets where image information can be highly utilized, such as sentiment analysis (SST-2) compared to tasks involving natural language inference (RTE). This suggests that our visual imagination-based approach can actually enhance the general language understanding capabilities by providing additional information.

\begin{table*}[ht!]
\centering
\resizebox{\textwidth}{!}{%
\begin{tabular}{@{}lccccccc@{}}
\toprule
\multicolumn{1}{l|}{Method} & \multicolumn{1}{c|}{KB} & $\alpha$NLI & CSQA & PIQA & SIQA & \multicolumn{1}{c|}{WG} & Avg. \\ \midrule
\multicolumn{8}{l}{\textbf{Pre-trained Language Models}} \\
\multicolumn{1}{l|}{GPT-2-L \citep{radford2019language-gpt2}} & \multicolumn{1}{c|}{-} & 56.5 & 41.4 & 68.9 & 44.6 & \multicolumn{1}{c|}{53.2} & 52.9 \\
\multicolumn{1}{l|}{RoBERTa-L \citep{DBLP:journals/corr/abs-1907-11692-roberta}} & \multicolumn{1}{c|}{-} & 65.6 & 45.0 & 67.6 & 47.3 & \multicolumn{1}{c|}{57.5} & 56.6 \\
\multicolumn{1}{l|}{DeBERTa-v3-L \citep{DBLP:conf/iclr/HeGC23-debertav3}} & \multicolumn{1}{c|}{-} & 59.9 & 25.4 & 44.8 & 47.8 & \multicolumn{1}{c|}{50.3} & 45.6 \\
\multicolumn{1}{l|}{Self-talk \citep{DBLP:conf/emnlp/ShwartzWBBC20-selftalk}} & \multicolumn{1}{c|}{-} & - & 32.4 & 70.2 & 46.2 & \multicolumn{1}{c|}{54.7} & - \\
\multicolumn{1}{l|}{COMET-DynGen \citep{DBLP:conf/aaai/BosselutBC21-dyngen}} & \multicolumn{1}{c|}{AT} & - & - & - & 50.1 & \multicolumn{1}{c|}{-} & - \\
\multicolumn{1}{l|}{SMLM \citep{DBLP:conf/emnlp/BanerjeeB20-smlm}} & \multicolumn{1}{c|}{*} & 65.3 & 38.8 & - & 48.5 & \multicolumn{1}{c|}{*} & - \\
\multicolumn{1}{l|}{GPT-2-L (MR; \citealp{DBLP:conf/aaai/MaIFBNO21-ma})} & \multicolumn{1}{c|}{AT} & 59.2 & 48.0 & 67.5 & 53.6 & \multicolumn{1}{c|}{54.7} & 56.6 \\
\multicolumn{1}{l|}{RoBERTa-L (MR; \citealp{DBLP:conf/aaai/MaIFBNO21-ma})} & \multicolumn{1}{c|}{AT} & 70.8 & 64.2 & 72.1 & 63.1 & \multicolumn{1}{c|}{59.6} & 66.0 \\
\multicolumn{1}{l|}{DeBERTa-v3-L (MR; \citealp{DBLP:conf/aaai/MaIFBNO21-ma})} & \multicolumn{1}{c|}{AT} & 76.0 & 67.0 & 78.0 & 62.1 & \multicolumn{1}{c|}{76.0} & 71.8 \\
\multicolumn{1}{l|}{MICO \citep{DBLP:conf/emnlp/SuWFZSZ22-mico}} & \multicolumn{1}{c|}{AT} & - & 44.2 & - & 56.0 & \multicolumn{1}{c|}{-} & - \\
\multicolumn{1}{l|}{Zero-shot Fusion \citep{DBLP:conf/naacl/KimKKAHY22}} & \multicolumn{1}{c|}{AT, CN, WD, WN} & 72.5 & 68.2 & 72.9 & 66.6 & \multicolumn{1}{c|}{60.8} & 68.2 \\
\multicolumn{1}{l|}{Multi-hop Knowledge Injection \citep{DBLP:journals/corr/abs-2305-05936-multi-hop}} & \multicolumn{1}{c|}{AT, CN, WD, WN} & 72.5 & 71.0 & 73.1 & - & \multicolumn{1}{c|}{61.0} & - \\
\multicolumn{1}{l|}{CAR-GPT-2-L \citep{DBLP:conf/emnlp/WangF0XLSB23-car}} & \multicolumn{1}{c|}{AbsAT} & 61.7 & 50.0 & 68.2 & 52.3 & \multicolumn{1}{c|}{55.2} & 57.5 \\
\multicolumn{1}{l|}{CAR-RoBERTa-L \citep{DBLP:conf/emnlp/WangF0XLSB23-car}} & \multicolumn{1}{c|}{AbsAT} & 72.7 & 66.3 & 73.2 & 64.0 & \multicolumn{1}{c|}{62.0} & 67.6 \\
\multicolumn{1}{l|}{CAR-DeBERTa-v3-L \citep{DBLP:conf/emnlp/WangF0XLSB23-car}} & \multicolumn{1}{c|}{AbsAT} & 79.6 & 69.3 & 78.6 & 64.0 & \multicolumn{1}{c|}{{\ul 78.2}} & 73.9 \\
\multicolumn{1}{l|}{CANDLE-DeBERTa-v3-L \citep{DBLP:journals/corr/abs-2401-07286-candle}} & \multicolumn{1}{c|}{CANDLE} & 81.2 & 69.9 & 80.3 & 65.9 & \multicolumn{1}{c|}{\textbf{78.3}} & {\ul 75.1} \\ \midrule
\multicolumn{8}{l}{\textbf{Large Language Models}} \\
\multicolumn{1}{l|}{GPT-3.5 (\texttt{text-davinci-003})} & \multicolumn{1}{c|}{-} & 61.8 & 68.9 & 67.8 & {\ul 68.0} & \multicolumn{1}{c|}{60.7} & 65.4 \\
\multicolumn{1}{l|}{ChatGPT (\texttt{gpt-3.5-turbo})} & \multicolumn{1}{c|}{-} & 73.2 & \textbf{75.7} & 81.7 & \textbf{69.7} & \multicolumn{1}{c|}{64.1} & 72.9 \\
\multicolumn{1}{l|}{GPT-4 (\texttt{gpt-4})} & \multicolumn{1}{c|}{-} & 75.0 & 43.0 & 73.0 & 57.0 & \multicolumn{1}{c|}{77.0} & 65.0 \\
\multicolumn{1}{l|}{LLAMA2-13B \citep{DBLP:journals/corr/abs-2307-092880-llama2}} & \multicolumn{1}{c|}{-} & 55.9 & 67.3 & 80.2 & 50.3 & \multicolumn{1}{c|}{72.8} & 65.3 \\
\multicolumn{1}{l|}{Mistral-v0.1-7B \citep{DBLP:journals/corr/abs-2310-06825-mistral}} & \multicolumn{1}{c|}{-} & 51.0 & 59.6 & \textbf{83.0} & 42.9 & \multicolumn{1}{c|}{75.3} & 62.4 \\
\multicolumn{1}{l|}{VERA-T5-xxl \cite{DBLP:conf/emnlp/0010WWS0H23-vera}} & \multicolumn{1}{c|}{AT} & 71.2 & 61.7 & 76.4 & 58.2 & \multicolumn{1}{c|}{67.2} & 66.9 \\
\multicolumn{1}{l|}{VERA-T5-xxl \citep{DBLP:conf/emnlp/0010WWS0H23-vera}} & \multicolumn{1}{c|}{AbsAT} & 73.2 & 63.0 & 77.2 & 58.1 & \multicolumn{1}{c|}{68.1} & 68.0 \\
\multicolumn{1}{l|}{CANDLE-VERA-T5-xxl \citep{DBLP:journals/corr/abs-2401-07286-candle}} & \multicolumn{1}{c|}{CANDLE} & 73.8 & 64.7 & 77.6 & 59.4 & \multicolumn{1}{c|}{71.3} & 69.4 \\ \midrule
\multicolumn{8}{l}{\textbf{Ours}} \\
\rowcolor{lightblue}
\multicolumn{1}{l|}{\textsc{Imagine}-GPT-2-L} & \multicolumn{1}{c|}{Synthetic VQA} & 61.5 & 63.9 & 68.9 & 53.0 & \multicolumn{1}{c|}{55.2} & 58.5 \\
\rowcolor{lightblue}
\multicolumn{1}{l|}{\textsc{Imagine}-RoBERTa-L} & \multicolumn{1}{c|}{Synthetic VQA} & 74.7 & 67.5 & 72.3 & 64.3 & \multicolumn{1}{c|}{61.2} & 68.0 \\
\rowcolor{lightblue}
\multicolumn{1}{l|}{\textsc{Imagine}-DeBERTa-v3-L} & \multicolumn{1}{c|}{Synthetic VQA} & \textbf{82.2} & {\ul 74.0} & {\ul 80.7} & 66.3 & \multicolumn{1}{c|}{76.7} & \textbf{76.0} \\ \midrule
\multicolumn{8}{l}{\textbf{Supervised \& Human}} \\
\multicolumn{1}{l|}{RoBERTa-L (Supervised)} & \multicolumn{1}{c|}{-} & 85.6 & 78.5 & 79.2 & 76.6 & \multicolumn{1}{c|}{79.3} & 79.8 \\
\multicolumn{1}{l|}{DeBERTa-v3-L (Supervised)} & \multicolumn{1}{c|}{-} & 89.0 & 82.1 & 84.5 & 80.1 & \multicolumn{1}{c|}{84.1} & 84.0 \\ 
\multicolumn{1}{l|}{Human} & \multicolumn{1}{c|}{-} & 91.4 & 88.9 & 94.9 & 86.9 & \multicolumn{1}{c|}{94.1} & 91.2 \\ \bottomrule
\end{tabular}%
}
\caption{Zero-shot evaluation results on five commonsense reasoning tasks (Accuracy \%). \textbf{Bold} and 
{\ul Underline} indicate the best and second-best results, respectively. AT, CN, WD, WN, and AbsAT refer to ATOMIC, ConcetNet, WikiData, WordNet, and AbstractATOMIC. The results of the large language models including GPT series are taken from \citet{DBLP:journals/corr/abs-2401-07286-candle}. SMLM (*) used different KBs for the different benchmarks.}
\label{tab:app_csqas}
\end{table*}

\end{document}